\documentclass{article}

\usepackage{nips_2018}

\usepackage[utf8]{inputenc} 
\usepackage[T1]{fontenc}    
\usepackage{hyperref}       
\usepackage{url}            
\usepackage{booktabs}       
\usepackage{amsfonts}       
\usepackage{nicefrac}       
\usepackage{microtype}      

\usepackage{graphicx}
\DeclareGraphicsExtensions{.pdf,.png,.jpg}
\usepackage{algorithm,algpseudocode}

\usepackage{subfigure}
\usepackage[letterpaper]{geometry}

\title{Regional Multi-scale Approach for Visually Pleasing Explanations of Deep Neural Networks}

\author{
  $^1$Dasom Seo, $^2$Kanghan Oh, $^1$Il-Seok Oh \\
  Division of Computer Science and Engineering\\
  Chonbuk National University \\
  $^1$\texttt{$\{$dasomseo, isoh$\}$@jbnu.ac.kr} \, $^2$\texttt{blastps@gmail.com} \\
}

\begin{document}

\maketitle

\begin{abstract}
  Recently, many methods to interpret and visualize deep neural network predictions have been proposed and significant progress has been made. However, a more class-discriminative and visually pleasing explanation is required. Thus, this paper proposes a \textit{region-based} approach that estimates feature importance in terms of appropriately segmented regions. By fusing the saliency maps generated from \textit{multi-scale} segmentations, a more class-discriminative and visually pleasing map is obtained. We incorporate this regional multi-scale concept into a prediction difference method that is model-agnostic. An input image is segmented in several scales using the super-pixel method, and exclusion of a region is simulated by sampling a normal distribution constructed using the boundary prior. The experimental results demonstrate that the regional multi-scale method produces much more class-discriminative and visually pleasing saliency maps.
\end{abstract}
\section{Introduction}
\label{intro}
Deep learning has facilitated breakthroughs for a variety of AI tasks and, in my cases, has achieved performance equal or superior to human performance [6]. Despite considerable success, most learning models do not satisfactorily explain why they reach a decision; thus, model deployment is delayed or even abandoned. Recognizing that this deficiency could cause potential harm, the European Union have adopted a regulation for algorithmic decision making, that addresses the “right to explanation” [5]. This regulation will restrict the deployment of AI systems that do not satisfy this constraint, which will have a significant impact on AI industry.

To address this situation, a variety of explanation techniques have been devised and evaluated [8]. The evaluation of some techniques involved user studies to ensure that the explanations increased user trust and helped users choose better models [9,12]. In the early days of development, interpreting the learning model itself dominated [4]. Currently, interpreting predictions for individual instances, which is the focus of this study, is receiving greater attention.

An interpretation model should satisfy several requirements. The most critical requirement is that the interpretation be class-discriminative, i.e., it should identify features that make the greatest contribution to determining the given class. As shown in Figure \ref{fig4}, sensitivity analysis (SA), gradient-weighted class activation map (Grad-CAM), and the proposed method are class-discriminative, whereas the other methods are not. This paper suggests another requirement, i.e., a visually pleasing saliency map. The proposed method demonstrates better performance for both requirements. In the first row of Figure \ref{fig4} where the target class is “eggnog,” note that the proposed method correctly indicates the eggnog regions inside the glass. In addition, our saliency map retains the object shapes clearly and thus visually pleasing. We believe that visually pleasing saliency maps can help machine learning experts select appropriate models and optimize hyper-parameters and are essential to help laypersons, such as doctors and public safety agents, choose and use a reasonable AI system. Our primary contribution is a \textit{region-based} approach that estimates feature importance in terms of appropriately segmented regions. By fusing saliency maps generated from multi-scale segmentations, a more class-discriminative and visually pleasing map is obtained. 

Avoiding a tradeoff between accuracy and interpretability is another important issue. For example, the class activation map (CAM) method working on only the convolutional neural networks (CNN) with global average pooling sacrifices $1\% \sim 5\%$ accuracy to obtain high class-discriminability [18]. In our implementation, we incorporate the regional multi-scale idea into a prediction difference method that is model-agnostic, i.e., applicable to any learning model without modifying internal operations.

The proposed method was evaluated qualitatively and quantitatively, and compared to state-of-the-art methods. The results demonstrate that the proposed method produces much more class-discriminative and visually pleasing saliency maps (Figure \ref{fig4}). In addition, the proposed method is two orders of magnitude faster than the conventional prediction difference algorithm.
\section{Related Works}
\label{rel}
Making a learning model interpretable involves interpreting the model itself and interpreting the predictions of individual input instances. The simplest form of interpreting a CNN model involves the visualization of filters and feature maps generated by the CNN, which includes an activation maximization technique that searches for an input-domain map that maximizes a node’s response [4], as well as schemes that are extended using effective regularizers [16].

The latter notion of instance interpretation seeks to explain why the model arrives at the classification decision given a particular instance and a target class. This paper focuses on this notion, particularly on methods that estimate the importance (or relevance) of features (pixels in case of images) with respect to a classification decision.

\textbf{Saliency maps explaining CNN decisions: }Given an instance image input to a trained CNN, the CNN generates a class prediction. Based on the prediction, an interpretation model measures the impact of a pixel or a region on the prediction. The value assigned to the corresponding pixel or region of the saliency map depends on the extent of the impact., i.e., greater impact results in a higher value. Several different methods to estimate the impact have been proposed.

Simonyan et al. proposed a gradient-based estimation method, which they referred to as SA [14]. The gradient indicates how much a small change in a pixel influences the class output. The gradient map itself is considered the saliency map. Zeiler and Fergus proposed a deconvolution method [17]. In their method, when a user designates a neuron in a hidden or output layer, the forward signal is reversed at the neuron and back-propagated to the input space. Bach et al. proposed the layer-wise relevance propagation (LRP) method in which class decision information in the output layer is decomposed into pixel-wise relevance values and back-propagated while satisfying the conservation rule [1]. Samek et al. emphasized the importance of quantitative evaluation and provided a rigorous comparison of the above-mentioned three methods [11]. Shrikumar et al. pointed out limitations with the gradient-based approach and proposed the DeepLIFT technique to address these problems [13]. Zhou et al. devised the CAM technique that exploits global average pooling [18]. Selvaraju et al. extended CAM to Grad-CAM to be applicable to a wider range of CNNs [12].

Explaining with text is also worthy of attention. Most studies employ a learning model comprising CNN and long short-term memory (LSTM) concatenation. Feature vectors output by the CNN encoder are passed to an LSTM-based decoder that generates explanatory texts. Barrat proposed a learning algorithm that uses training sample $(x,y,E)$ where $x$ and $y$ are the input image and class label, respectively [2]. Here, $E$ represents the class-discriminative textual features such as “this is a bird with a white belly, brown back, and a white eyebrow”. Dong et al. extended the text explanation to video clips [3].

Robnik-Sikonja et al. proposed the prediction difference method [10], and several studies have extended and improved this method [3,20]. The proposed regional and multi-scale approach is embedded in the prediction difference; thus, we review prediction difference in greater detail.

\textbf{Prediction difference: }Prediction difference measures the difference between prediction for a feature vector and prediction for the feature vector without feature $i$. The essential problem is how to exclude the feature $i$ from the original feature vector. Robnik-Sikonja et al. proposed a principled method that \textit{simulates} feature exclusion based on marginal probability [10]. A formal generic code will be presented in Section \ref{alg1}. Since excluding a pixel (feature) or a region is impossible in a CNN, a reasonable simulation method is required. Zintgraf et al. presented solutions that use prior knowledge about the image characteristics [20]. Then, they applied the solutions to a CNN that classifies natural images in ImageNet. Conditional sampling considers the fact that a given pixel value is highly dependent on neighboring pixels, and multivariate analysis excludes a rectangular region rather than a single pixel (Section \ref{alg2}).

Dong et al. improved the interpretability of a video captioning CNN-RNN model by embedding the prediction difference maximization operation in the model [3]. The learning stage uses an objective function that maximizes the discriminability and interpretability of a learning model. Interpretability is represented by a topic model (semantic representation) constructed by parsing target text descriptions. After completing the learning stage, prediction difference maximization is applied to search for correspondence between neurons (features extracted from video frames by a CNN) and semantic topic words. The correspondence information enables a neuron to visualize the activation levels for both relevant and irrelevant words across consecutive video frames.

\textbf{Comparative studies and discussions: }We can characterize and compare the above-mentioned methods in terms of several criteria. The proposed method corresponds to former choices for four criteria, i.e., regional, multi-scale, model-agnostic, and syntactic.

$\bullet$ Regional vs. non-regional: As shown in Figure \ref{fig4}, all methods except the proposed are pixel-wise. In addition, compared to the proposed method, their saliency maps are less class-discriminative and not visually pleasing.

$\bullet$ Multi-scale vs. single-scale: The regional approach requires segmenting an image into a set of regions. The granularity (size) of the segmented regions is very important. Multi-scale processing that fuses maps from coarse to fine scales is one of our key ideas. Pixel-wise methods can be considered as a single-scale processing that relies only on atomic scale.

$\bullet$ Model-agnostic vs. model-dependent: The methods based on the prediction difference are inherently model-agnostic, i.e., applicable to any learning models, because they rely only on the output values regardless of internal workings of the models. The local interpretable model-agnostic explanation (LIME) is model-agnostic because it optimizes a separate model $g$ to explain $f$ where $g$ should be interpretable such as linear classifier or decision tree [9]. The LIME is cumbersome since $g$ should be faithful to $f$; therefore, another model selection and optimization for $g$ should be solved properly. The other above mentioned methods modify the model’s internal operations or rely on the model’s internal values; thus, they are model-dependent.

$\bullet$ Syntactic vs. semantic: The saliency maps in Figure \ref{fig4} show the activation levels to explain the extent to which parts of the image influence the classification decision. Thus, the maps are considered as interpreting CNN decisions syntactically. In contrast, textual descriptions that include topic models are considered to provide semantic interpretations [2,3].

\section{Algorithms}
\label{Alg}
When explaining the classifier’s decision, the causal effect that the change of one feature value or the values of a feature subset has on the prediction value can be used as essential information. The greater the effect, the more important we consider the changed features. The principle of prediction difference originates from this simple observation. Specifically, the prediction difference measures the value $f(\textbf{x})-f(\textbf{x}_{\setminus i})$ between the prediction $f(\textbf{x})$ with all features and the prediction $f(\textbf{x}_{\setminus i})$ with all features except the $i$-th feature. The feature vector \textbf{x} is a \textit{d}-dimensional vector $\textbf{x}=(x_1,x_2,\cdots, x_d )$. In practice, the difference is measured relative to a specific class label, and the classifiers provide a probability for each of the class labels. Thus, the term $f(\textbf{x})$ is replaced with $p(y|\textbf{x})$ in the following descriptions.
\subsection{Generic code of prediction difference}
\begin{algorithm}
\caption{Generic Code of Prediction Difference}\label{alg:rmleft}
\begin{algorithmic}[1]
\Require{trained classifier $f$, instance feature vector \textbf{x}, target class $c$}
\Ensure{saliency vector \textbf{s}}
\State Run $f$ to get $p(y=c|\textbf{x})$
\For {each feature $i$ of \textbf{x}}
    \State Estimate $p(y=c|\textbf{x}_{\setminus i})$ using (1)
    \State $s_{i} = g \left ( p(y=c|\textbf{x}),p(y=c|\textbf{x}_{\setminus i}) \right)$
    \EndFor
\end{algorithmic}
\label{ag1}
\end{algorithm}
The procedure of the prediction difference is described by the generic code of Algorithm \ref{ag1} [10].The code is a type of meta-algorithm where several options should be specified during implementation, such as classifier type and method for excluding feature $i$. Robnik-Sikonja et al. proposed a simulation scheme that formulates the exclusion of feature $i$ by marginalization in Eq. (\ref{eq1}), where the $i$-th feature has the domain $x_i \in \{ {a_1,a_2,\cdots ,a_{m_i} }\}$. Since the term $p(x_i=a_j\mid \textbf{x}_{\setminus i})$ in the first summation is in most cases impossible to calculate, the second summation is used as an approximation. Here, $\textbf{x} \leftarrow x_i=a_j$ denotes the feature vector \textbf{x} where the value of $x_i$ is set to be $a_j$.
\begin{equation}
    p(y|\textbf{x}_{\setminus i}) = \sum_{j=1}^{m_i}p(x_i=a_j \mid \textbf{x}_{\setminus i})p(y\mid \textbf{x} \leftarrow x_i=a_j) \cong \sum_{j=1}^{m_i}p(x_i=a_j)p(y\mid \textbf{x} \leftarrow x_i=a_j)
    \label{eq1}
\end{equation}

In Algorithm \ref{ag1}, $g$ in line 4 is a function that computes the prediction difference. The simplest function is the subtraction $g(a,b)=a-b$. Refer to [10] for other schemes relative to information difference and the weight of evidence.
    \label{alg1}
\subsection{Adapting to images}
    \label{alg2}
\begin{figure}
  \centering
  \subfigure[]{
   \includegraphics[scale=0.85]{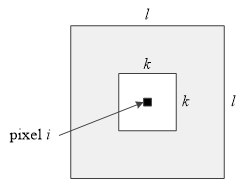}
  }
  \subfigure[]{
  \includegraphics[scale=0.9]{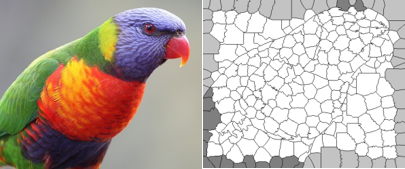}
  }
  \caption{Gray regions used to estimate normal distribution (a) Conditional sampling and multivariate analysis [20] (b) Regional multi-scale method (Proposed)}
  \label{fig1}
\end{figure}

\begin{algorithm}
\caption{Conditional Sampling and Multivariate Analysis for Prediction Difference}\label{alg:rmleft}
\begin{algorithmic}[1]
\Require{trained classifier $f$, image \textbf{x}, target class $c$, sampling number $r$, patch sizes $k$ and $l(l>k)$}
\Ensure{saliency map \textbf{s}}
\State \textbf{c}=0 // having the same size as \textbf{x}
\State Run $f$ to get $p(y= c\mid \textbf{x})$
\For {each pixel $i$ of \textbf{x}}
    \State $\textbf{x}'$ = \textbf{x}; $sum=0$
    \State Define $k\times k$ $\textbf{x}_{in}$ and $l\times l$ $\textbf{x}_{out}$ patches centered at $i$ // see Figure \ref{fig1}(a)
    \For{$j$=1 to $r$} // sampling $r$ times
        \State Replace $\textbf{x}_{in}$ of $\textbf{x}'$ with a patch sampled from $p(\textbf{x}_{in} \mid \textbf{x}_{out \setminus \textbf{x}_{in}})$
        \State Run $f$ to get $p(y=c \mid \textbf{x}')$
        \State $sum \mathrel{+}= p(y=c\mid \textbf{x}')$
    \EndFor
    \State $p(y=c\mid \textbf{x}_{\setminus i}) = sum/r$
    \For{every pixel $q$ in $\textbf{x}_{in}$}
        \State $s_q=g \left ( p(y=c|\textbf{x}),p(y=c|\textbf{x}_{\setminus i}) \right); c_q++$
    \EndFor
\EndFor
\State \textbf{s}=\textbf{s}/\textbf{c} // element-wise division
\end{algorithmic}
\label{ag2}
\end{algorithm}
Zintgraf et al. extended Algorithm \ref{ag1} to processing images with a CNN [20]. In Algorithm \ref{ag2}, line 7 requires further explanation. The goal of this line is to accomplish exclusion of $\textbf{x}_{in}$ from $\textbf{x}'$, i.e., $\textbf{x}'_{\setminus\textbf{x}_{in}}$. Algorithm \ref{ag2} simulates this by replacing $\textbf{x}_{in}$ of $\textbf{x}'$ with a sampled patch. The sampling is performed from a distribution of $p(\textbf{x}_{in} |\textbf{x}_{out\setminus\textbf{x}_{in}})$. In the implementation, the mean and variance of a normal distribution are calculated from the pixels in $\textbf{x}_{out\setminus\textbf{x}_{in}}$, i.e., the gray area in Figure \ref{fig1}(a). Line 7 substitutes the values of the pixels in the patch $\textbf{x}_{in}$, i.e., the white area of Figure \ref{fig1}(a), with values obtained by sampling the normal distribution. In addition, to simulate the marginal probability, Algorithm \ref{ag2} executes line 7 $r$ times, and line 11 takes the average. Note that the average is assigned to every pixel in the patch $\textbf{x}_{in}$ by line 13 to obtain a smooth saliency map.

Example saliency maps produced by Algorithm \ref{ag2} are shown in Figure \ref{fig4}(d). Though the algorithm attempts multiple sampling and propagation of the average to neighboring pixels, the output saliency map is still problematic in the sense that saliency values are dispersed inside object regions and leaked into the background regions. As a result, such a saliency map does not appear to be particularly useful when explaining CNN predictions.

The computationally expensive step is line 8, which performs forward computation, and Algorithm \ref{ag2} is very slow because this line is executed $rn^2$ times for an $n\times n$ image. Algorithm \ref{ag2} required 70 minutes on a VGG network with a GPU for a $256\times 256$ image where $k=10$, $l=14$, and $r=10$ [20].
\subsection{Regional multi-scale prediction difference}
\label{alg3}
If we continue to employ pixel-wise processing, the dispersion and leakage problems of Algorithm \ref{ag2} cannot be remedied. Thus, as a solution, we propose a \textit{regional} approach that is easily embedded into the generic code of Algorithm \ref{ag1}. The proper granularity of regions is not known in advance; therefore, we also propose a multi-scale approach. These regional and multi-scale approaches are combined and described in Algorithm \ref{ag3}.
\begin{algorithm}
\caption{: Regional Multi-scale Prediction Difference}\label{alg:rmleft}
\begin{algorithmic}[1]
\Require{trained classifier $f$, image \textbf{x}, target class $c$, multi-scale number $r$}
\Ensure{saliency map \textbf{s}}
\State Run $f$ to get $p(y=c\mid \textbf{x})$
\State Estimate $N(\mu, \sigma^2)$ from the image \textbf{x} segmented with scale $2^8$
\For{$j=1$ to $r$} // for each scale
    \State Perform a segmentation with the scale $2^j$ // super-pixel segmentation
    \For{each region $i$}
        \State Simulate the exclusion of $i$, i.e., $\textbf{x}_{\setminus i}$
        \State Run $f$ to get $p(y=c\mid \textbf{x}_{\setminus i})$
        \For{each pixel $q$ in the region $i$} 
            \State$s^{j}_{q}=max \left ( 0, g\left (p(y=c|\textbf{x}),p(y=c|\textbf{x}_{\setminus i}) \right) \right)$
        \EndFor
    \EndFor
\EndFor
\State $\textbf{s} = \frac{1}{r}\sum_{j=1}^{r}\textbf{s}^j$
\end{algorithmic}
\label{ag3}
\end{algorithm}

In Algorithm \ref{ag3}, lines 2 and 4 require the region segmentation. The super-pixel method has the advantage that the number of resulting regions is controllable, thereby enabling multi-scale processing; therefore, we adopt a previously proposed algorithm [7]. Note that line 2 uses a fixed scale $2^8$ and line 4 goes through multiple scales ($2^1$ to $2^r$). In addition, lines 2 and 6 cooperate to simulate region exclusion, and these lines exploit the boundary prior, i.e., the inherent nature of boundary regions belonging primarily to backgrounds [19]. Line 2 prepares a normal distribution that will be used by line 6 for sampling background-like pixel values. Here, the RGB mean vector for each of the boundary regions (shown in Figure \ref{fig1}(b) as gray super-pixels) is calculated separately, and those mean vectors are used as sample points to estimate the normal distribution. Note that some object regions (e.g., those shown as dark gray super-pixels in Figure \ref{fig1}(b)) may be boundary regions; thus, to avoid them from participating in the estimation, we apply the mean shift algorithm and select the largest cluster, where the mode of the largest cluster is taken as the mean $\mu$ of the normal distribution and the variance $\sigma^2$ is fixed at 10.0. Line 6 simulates the exclusion of region $i$. Here, each pixel in the region is substituted with a sample drawn from $N(\mu,\sigma^2)$. Note that the negative impact is ignored using the max operator in line 9.

Line 7 executes the forward computation of a deep neural network. Here, the number of executions is $2^1+2^2+\cdots+2^r$, e.g., 62 executions for $r=5$. Compared to $rn^2$ in Algorithm \ref{ag2}, this represents dramatically lower computational complexity.
\section{Experimental Results}
\label{exper}
\begin{figure}
  \centering
  \includegraphics[scale=0.5]{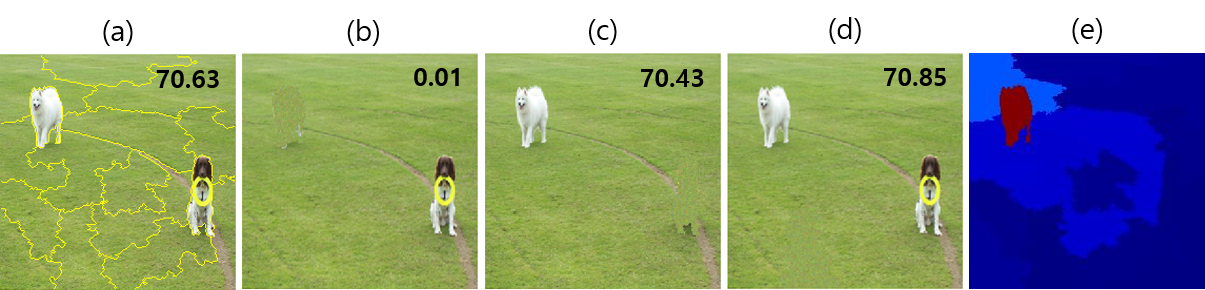}
  \caption{Effect of regions foe “Samoyed” class image from ImageNet (number at top-right corner is the prediction score for the images; multi-scale fusion is shown in Figure \ref{fig4}) (a) input image overlaid with 16-level segmentation information where each of the two dogs corresponds to a single super-pixel; (b) Samoyed (target class) is erased; (c) other dog (not target class) is erased; (d) grass is erased; (e) saliency map.}
  \label{fig2}
\end{figure}
\begin{figure}
  \centering
  \includegraphics[scale=0.4]{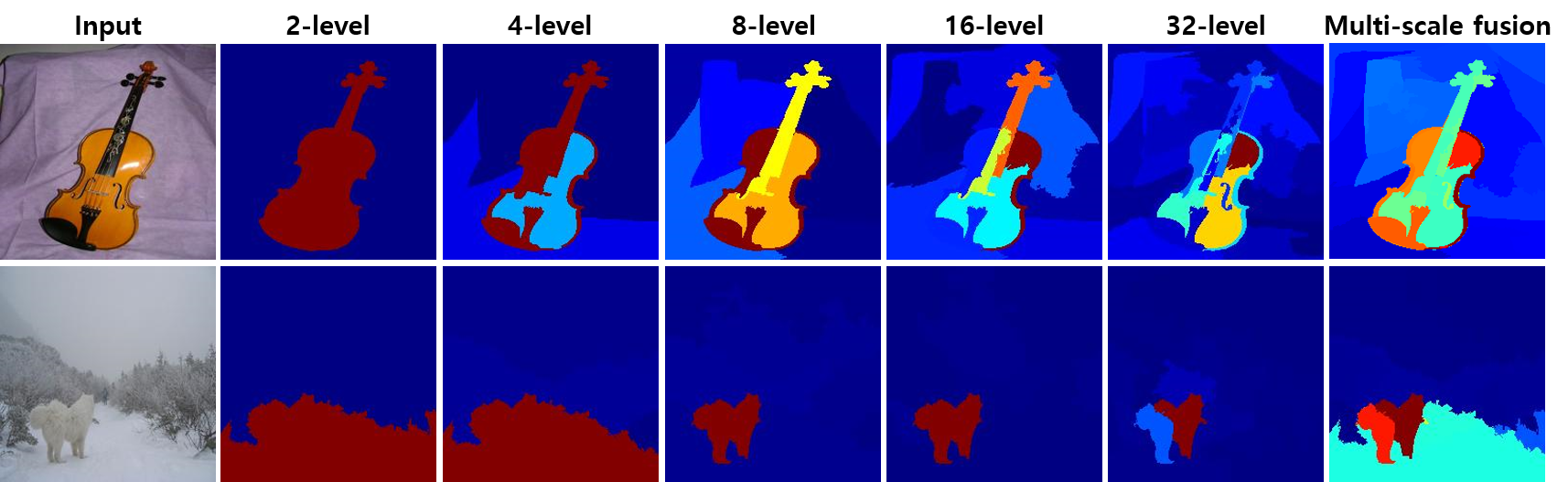}
  \caption{Saliency maps from coarse to fine-scale for two example images chosen from “violin” and “Samoyed” classes (the maps are heavily influenced by segmentation scales).}
  \label{fig3}
\end{figure}
\begin{figure}
  \centering
`\includegraphics[scale=0.42]{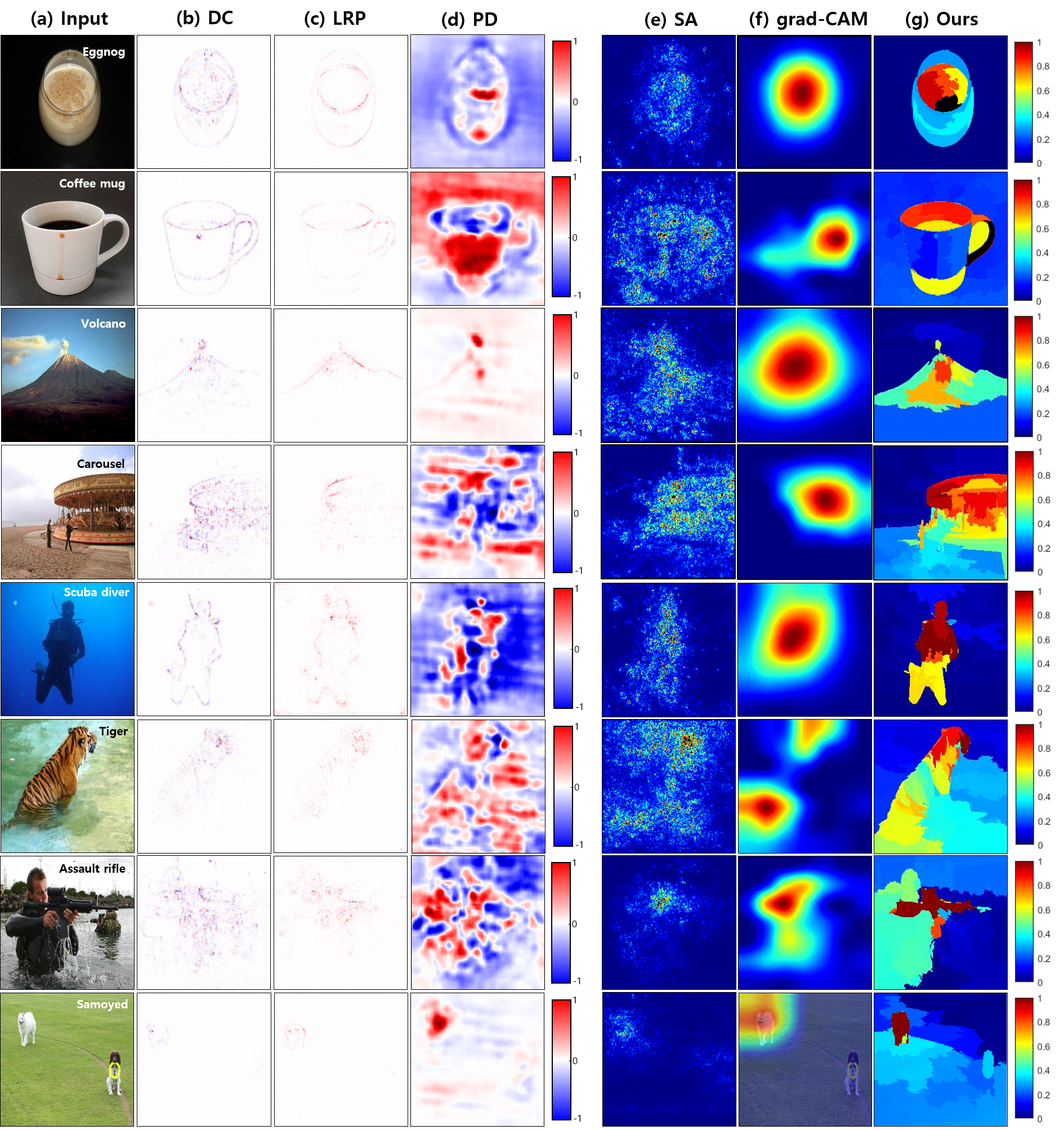}
  \caption{Visual comparison of conventional and proposed methods using GoogLeNet. The conventional methods are deconvolution (DC) [17], sensitivity analysis (SA) [14], layer-wise relevance propagation (LRP) [1], prediction difference in Algorithm \ref{ag2} (PD) [20], and gradient-weighted class activation mapping (Grad-CAM) [12]. Note two color bars representing the ranges. The maps of the five conventional methods were made using official software (\url{https://github.com/lmzintgraf/DeepVis-PredDiff}) and an official site (\url{ https://lrpserver.hhi.fraunhofer.de/image-classification}).}
  \label{fig4}
\end{figure}
In our experiments, the visual and quantitative comparisons of five state-of-the-arts methods and the proposed method (Algorithm \ref{ag3}) were conducted on the ILSVRC 2012 dataset (ImageNet). For the trained model, we used the GoogLeNet [15]. We set the input parameter $r$ of Algorithm \ref{ag3} to 5. The experiments were performed on an Intel Core i5-4670 CPU (3.40 GHz) with a Nvidia GTX1080 GPU. The average computation time of Algorithm \ref{ag3} was approximately 5 s, which is two orders of magnitude faster than Algorithm \ref{ag2} [20].

\textbf{Effect of regions: }
As the most important analysis of our algorithm, Figure \ref{fig2} demonstrates the causal effect the exclusion of a region has on the prediction score. When the region corresponding to the target class “Samoyed” is erased, as shown in Figure \ref{fig2}(b), the prediction score dramatically drops from 70.63\% to 0.01\%. The other cases of erasing irrelevant regions shown in Figure \ref{fig2}(c)-(d) show little effect. We assure that the CNN can distinguish different species of dogs because no impact is observed when erasing other dog. Figure \ref{fig2}(e) shows the resulting saliency map for the scale of 16-level segmentation. From these observations, we believe that the proposed regional approach with super-pixel segmentation and boundary prior-based region exclusion simulation is very effective and viable.

The second analysis is related to the multi-scale processing. The objects in natural images appear in various sizes. Figure \ref{fig3} shows two example images where the violin is larger than the Samoyed. Each row shows the saliency maps resulting from 2- to 32-level segmentations. As shown in the top row, coarse-scale segmentation well explains the global shape of the violin, while fine-scale segmentation explains the details faithfully. In the bottom row, the target object is well explained in middle (8-level) to fine scales (16- and 32-level).
\begin{figure}
    \centering
    \includegraphics[scale=0.35]{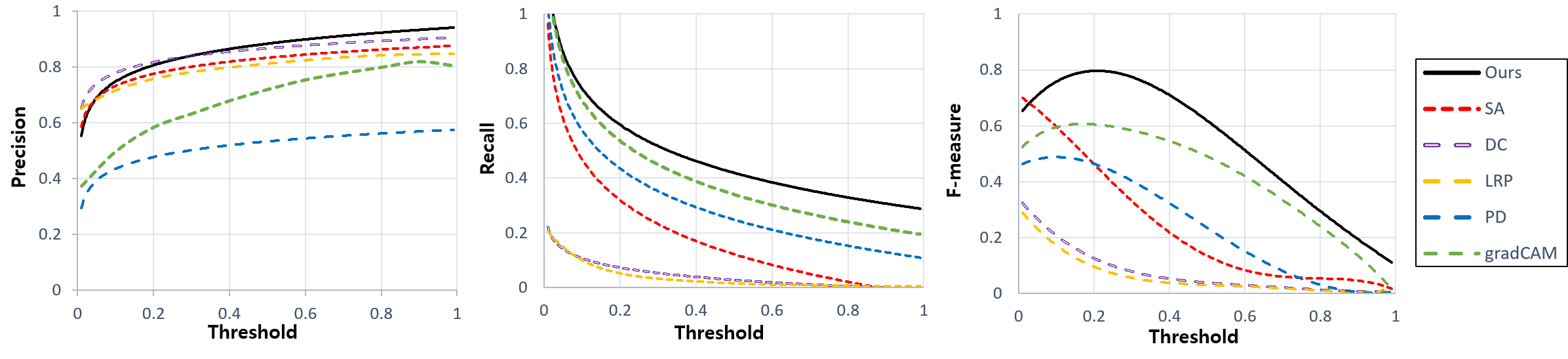}
    \caption{Quantitative comparison of saliency maps using precision, recall, and F-measure curves.}
    \label{fig5}
\end{figure}

\textbf{Comparison of visual quality: }Figure \ref{fig4} compares the results of the proposed method to those of five conventional state-of-the-art methods. The saliency maps of DC, LRP, and SA look like point clouds, and those of PD and Grad-CAM are similar to blobs and contours, respectively. The proposed method produces saliency maps that retain the object shapes effectively, leading to a visually pleasing property. Does this visually pleasing really matter? A user can easily perceive the heated objects in a saliency map produced by the proposed method; thus, it is expected that the user will trust the system more. In the saliency maps produced by the Grad-CAM method, the user must attempt to identify objects by looking at the original image and saliency map alternately. The bottom row of Figure \ref{fig4} attempts to overlap the saliency map of the Grad-CAM method onto the original image; however, in the resulting image, it is difficult to perceive the heated object. We believe that the visually pleasing property of the proposed method leads to perceptual attractiveness and increases user trust in the system.

In terms of class-discriminability, the DC and LRP methods reveal a clear limitation, i.e., they tend to emphasize object edges and do not hit the right places for the target class. For example, the images belonging to the “eggnog” and “coffee mug” classes should be differently hit on the inside content and container surface, respectively. However, the DC and LRP methods only found the edges of the container for both classes. This observation indicates that these methods cannot distinguish the eggnog and coffee mug. The Grad-CAM method correctly hit the inside content and the handle of the container for these two classes, respectively. Note that these are also well distinguished with the proposed method. The SA and PD are not clear whether they distinguish two classes or not. In consideration of the qualities of the other images in Figure \ref{fig4}, we conclude that the proposed and Grad-CAM methods show greater class-discriminative ability than the other compared methods.

We have experimented with many other images in ImageNet and other datasets and carefully analyzed their saliency maps. As a result, we conclude that the observational facts disscussed so far hold consistently. Experiments with AlexNet, VGG-16, and ResNet-101 produced similar performance as GoogLeNet.

\textbf{Quantitative comparisons: }Note that a metric to evaluate the quality of saliency maps has not been well established. The area over the MoRF perturbation curve has been proposed and used to compare SA, DC, and LRP quantitatively [11]; however, it is not intuitive. As an alternative, we adopt the evaluation protocol used to evaluate object localization methods. To simplify the evaluation, we selected 100 images from ImageNet that are composed of only an object and white background. 

The standard metrics, i.e., precision, recall, and F-measure, are measured by matching the ground truth and binary map obtained by thresholding a saliency map. Figure \ref{fig5} shows three curves drawn by differing the threshold from 0 to 1. Overall, the proposed method is superior to other methods in both precision and recall. The Grad-CAM method ranks second in terms of F-measure. The F-measure of SA is high until the threshold becomes approximately 0.2; however, this value is then reduced rapidly due to its low recall. The DC and LRP methods maintain high precision while demonstrating low overall recall. Their low recall is primarily attributed marking object boundaries.

\section{Conclusion}
\label{conc}
This paper has proposed a new requirement, i.e., being visually pleasing, that is important for interpreting the prediction in a perceptually attractive manner. An intelligent system that enable a visually pleasing interpretation is expected to receive higher trust from users. As a viable solution, this paper has presented the regional multi-scale prediction difference method. Through comprehensive experimental analysis, we have demonstrated that the proposed method is much more class-discriminative and visually pleasing than state-of-the-art methods. In future, we plan to perform actual user studies. Embedding the regional multi-scale concept to other methods, such as Grad-CAM, will be another focus of future work. In addition, semantic interpretation by incorporating our method into CNN-RNN models will also be considered in future.
\section*{References}
\small
[1] Bach, S., Binder, A., Montavon, G., Klauschen, F., Müller, K. R., \& Samek, W. (2015). On pixel-wise explanations for non-linear classifier decisions by layer-wise relevance propagation. {\it PloS One}, 10(7), e0130140.

[2] Barratt, S. (2017). InterpNET: neural introspection for interpretable deep learning. In {\it Symposium on Interpretable Machine Learning}.

[3] Dong, Y., Su, H., Zhu, J., \& Zhang, B. (2017). Improving interpretability of deep neural networks with semantic information. {\it arXiv preprint arXiv:1703.04096}.

[4] Erhan, D., Bengio, Y., Courville, A., \& Vincent, P. (2009). Visualizing higher-layer features of a deep network. {\it University of Montreal}, 1341(3), 1. 

[5] Goodman, B., \& Flaxman, S. (2016). European Union regulations on algorithmic decision-making and a" right to explanation". {\it arXiv preprint arXiv:1606.08813}.

[6] LeCun, Y., Bengio, Y., \& Hinton, G. (2015). Deep learning. {\it Nature}, 521(7553), 436.

[7] Liu, M. Y., Tuzel, O., Ramalingam, S., \& Chellappa, R. (2011, June). Entropy rate superpixel segmentation. In {\it IEEE Conference on Computer Vision and Pattern Recognition}, pp. 2097-2104.

[8] Montavon, G., Samek, W., \& Müller, K. R. (2017). Methods for interpreting and understanding deep neural networks. {\it Digital Signal Processing}, pp.1-15. 

[9] Ribeiro, M. T., Singh, S., \& Guestrin, C. (2016, August). Why should i trust you?: Explaining the predictions of any classifier. In {\it The 22nd ACM SIGKDD International Conference on Knowledge Discovery and Data Mining}, pp. 1135-1144. 

[10] Robnik-Šikonja, M., \& Kononenko, I. (2008). Explaining classifications for individual instances. {\it IEEE Transactions on Knowledge and Data Engineering}, 20(5), 589-600. 

[11] Samek, W., Binder, A., Montavon, G., Lapuschkin, S., \& Müller, K. R. (2017). Evaluating the visualization of what a deep neural network has learned. {\it IEEE Transactions on Neural Networks and Learning Systems}, 28(11), 2660-2673.

[12] Selvaraju, R. R., Cogswell, M., Das, A., Vedantam, R., Parikh, D. \& Batra, D. (2017). Grad-CAM: Visual explanations from deep networks via gradient-based localization. In {\it International Conference on Computer Vision}, pp. 618-626.

[13] Shrikumar, A., Greenside, P., \& Kundaje, A. (2017). Learning important features through propagating activation differences. {\it arXiv preprint arXiv:1704.02685}.

[14] Simonyan, K., Vedaldi, A. \& Zisserman, A. (2013). Deep inside convolutional networks: visualising image classification models and saliency maps. In {\it International Conference on Learning Representations Workshop}.

[15] Szegedy, C., Liu, W., Jia, Y., Sermanet, P., Reed, S., \& Anguelov, D. \& Rabinovich, A.(2015). Going deeper with convolutions. In {\it IEEE Conference on Computer Vision and Pattern Recognition}, pp. 1-9.

[16] Yosinski, J., Clune, J., Nguyen, A., Fuchs, T., \& Lipson, H. (2015). Understanding neural networks through deep visualization. In {\it International Conference on Machine Learning}

[17] Zeiler, M. D., \& Fergus, R. (2014, September). Visualizing and understanding convolutional networks. In {\it European Conference on Computer Vision}, pp. 818-833.

[18] Zhou, B., Khosla, A., Lapedriza, A., Oliva, A., \& Torralba, A. (2016, June). Learning deep features for discriminative localization. In {\it IEEE Conference on Computer Vision and Pattern Recognition}, pp. 2921-2929.

[19] Zhu, W., Liang, S., Wei, Y., \& Sun, J. (2014). Saliency optimization from robust background detection. In {\it IEEE Conference on Computer Vision and Pattern Recognition}, pp. 2814-2821.

[20] Zintgraf, L. M., Cohen, T. S., Adel, T., \& Welling, M. (2017). Visualizing deep neural network decisions: Prediction difference analysis. In {\it International Conference on Learning Representations}.

\end{document}